\documentclass[journal,twoside,web]{ieeecolor}
\usepackage{generic}
\usepackage{cite}
\usepackage{amsmath,amssymb,amsfonts}
\usepackage{algorithm}
\usepackage{algorithmic}
\usepackage{graphicx}
\usepackage{textcomp}
\newcommand{\myreferences}{references}
\def\BibTeX{{\rm B\kern-.05em{\sc i\kern-.025em b}\kern-.08em
		T\kern-.1667em\lower.7ex\hbox{E}\kern-.125emX}}
\begin{document}
\title{Autonomous Situational Awareness for Robotic Swarms in High-Risk Environments}
\author{Vincent W. Hill, \IEEEmembership{Member, IEEE}, Ryan W. Thomas, \IEEEmembership{Member, IEEE}, and Jordan D. Larson, \IEEEmembership{Member, IEEE}
\thanks{The authors are with the Department of Aerospace Engineering and Mechanics, the University of Alabama, Tuscaloosa, AL 35401 USA. (e-mail: vwhill@crimson.ua.edu; rthomas8@crimson.ua.edu; jdlarson1@eng.ua.edu). }}

\maketitle

\begin{abstract}
This paper describes a technique for the autonomous mission planning of robotic swarms in high-risk environments where agent disablement is likely. Given a swarm operating in a known area, a central command system generates measurements from the swarm. If those measurements indicate changes to the mission situation such as target movement or agent loss, the swarm planning is updated to reflect the new situation and guidance updates are broadcast to the swarm. The primary algorithms featured in this work are A* pathfinding and the Generalized Labeled Multi-Bernoulli multi-object tracking method.
\end{abstract}


\section{Introduction}
The massive increase in sensor and computational power over the last decade has greatly accelerated the development of autonomous robots. The next step in this evolution is to team groups of robots together in a swarm to accomplish tasks that are either too complicated or dangerous for one vehicle to undertake. This adds a layer of complexity to the autonomous guidance, navigation, and control (GNC) problem, since now there is a team of agents that must act in concert and respond to changes in situation without human input.
 
The majority of research in the current literature involving multi-agent, multi-target GNC either assume perfect information or do not directly consider the possibility of randomly changing number of objects, missed detections, or cluttered measurements together. Pierson and Rus \cite{cluttrack} used a modified Voronai tessellation to ensure obstacle avoidance during target pursuit, but assumed perfect information of both obstacles and targets. Foderaro et. al. \cite{DOC} provided a technique for optimal control of a sensor network for target tracking. They consider both target motion and the possibility of missed detections, but do not account for potential agent loss mid-mission or clutter measurements. You et. al. \cite{coop_sense} use a recursive project algorithm to cooperatively localize a mobile sensor network. Despite using range measurements for localization, these authors assume perfect information in their approach. We propose a technique for autonomous swarm operations that takes into account considerations involving incomplete or significantly uncertain information, such as agent loss, cluttered measurements, missed detections, and target motion, neglected in most prior research through an approach using Random Finite Sets (RFS).

In a parallel field to GNC, multi-object tracking (MOT) problems have been studied extensively over the last two decades. One significant contribution is the development of MOT filters based on RFS, in particular work done by Mahler and the Vo brothers in developing the Gaussian Mixture Probability Hypothesis Density (GM-PHD) \cite{Vo_GMPHD} and eventually the Generalized Labeled Multi-Bernoulli (GLMB) \cite{ReuterLMB} filters. The core concept behind RFS filtering is to regard the collection of target states and measurements received as sets of both unknown values and number of elements (cardinality), the probability densities of which can then be propagated through Bayesian methods for finite set statistics \cite{Mahler_MTBayes}. This is beneficial when the number of objects is generally unknown and there is a chance of missed detections or spurious measurements not related to objects (clutter), which is generally the case during the operation of many real-world robotic swarms. In dynamic, high-risk environments friendly objects, now known as swarm agents, are likely to be incapacitated. Additionally, new agents can be introduced, mission objectives (referred to as targets in this work) can move, new obstacles can be discovered, and other features indistinct from agent-generated measurements are common. Hence the RFS formulation is natural to oversee swarm operations.

Some research has been accomplished regarding the control of large swarms using RFS theory. Doerr and Linares \cite{doerr_conf} defined the swarming formation problem as the distance between the current swarm RFS distribution and a desired distribution, then applied model predictive control to carry out the guidance commands. A subsequent work by these researchers \cite{doerr_arxiv} on the swarm control problem used sparse RFS control gain matrices, allowing agents to use only local or personal information topology to drive control. However, these works do not consider moving targets or operations in a high-risk environment.

The authors have developed several techniques that extend the RFS multi-target tracking filters to guidance, navigation, and planning \cite{JDL_Linares_ION} \cite{JDL_Thomas_ELQR} \cite{hicard}. This work will build upon these by incorporating dynamic planning and guidance updating based on target movement and agent death, a GLMB filter for the swarm state estimation, and expanded results from the conference paper \cite{aero2021}. Specifically this work explicitly considers the possibility of mid-mission agent death, which occurs via a seeded random process.

The core procedures of this work are as follows. A human mission initiator defines the mission area size, initial agent and target locations, and obstacles, if any. A mission plan for each agent and the swarm as a whole must be autonomously devised and transmitted to the swarm. This is accomplished by a coupled A*/Hungarian algorithm Simultaneous Target Assignment and Trajectory (STAT) optimization technique. To accomplish the A* trajectory optimization, the mission area is discretized and all possible paths through the discrete nodes between the agent and target are checked until a path with the lowest possible cost is found. After the A* trajectories for each agent-target assignment is determined, the Hungarian algorithm determines the optimal assignments based on total distance traveled by the team. Next, a set of waypoints from each agent's starting location to its target is mapped from the A* grid to the physical mission area and downlinked to each agent's local computer. Each agent must generate its own control law to stabilize itself and carry out its associated guidance command, recognize when it reaches each waypoint, and adjust its state to reach the next waypoint. Figure 1 provides a general overview of the approach.

\begin{figure}\label{OneColumn}
	\centering
	\includegraphics[width=3.25in]{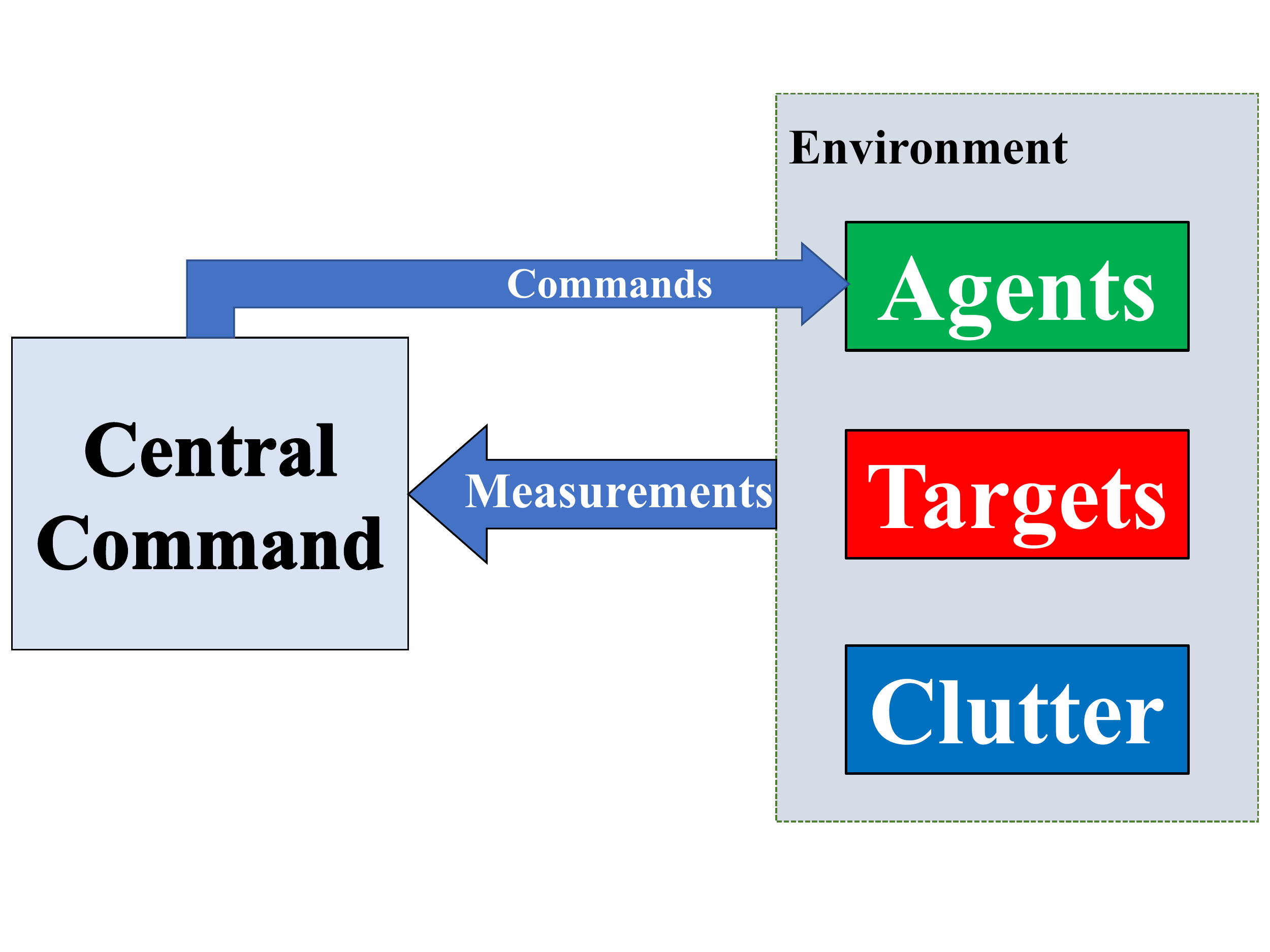}\\
	\caption{\textbf{Problem approach flowchart}}
\end{figure}

A central command system, which could be envisioned as a high altitude aircraft with a sensor suite and a powerful processor, must track the swarm as it carries out its mission. If the operation is occurring in a dynamic, high-risk environment, some agents are likely to be incapacitated and targets are likely to move. Central command must recognize when this occurs and adjust the swarm guidance to ensure mission completion. This is accomplished purely through measurements, eliminating the need for two-way communications in this application. A drawing describing an example scenario is given in Figure 2, where the aircraft is the central command, green quadrotors are agents, red quadrotors are targets, and the pillars with falling rocks represent a dangerous, cluttered environment with obstacles.

\begin{figure}\label{OneColumn}
	\centering
	\includegraphics[width=2.0in]{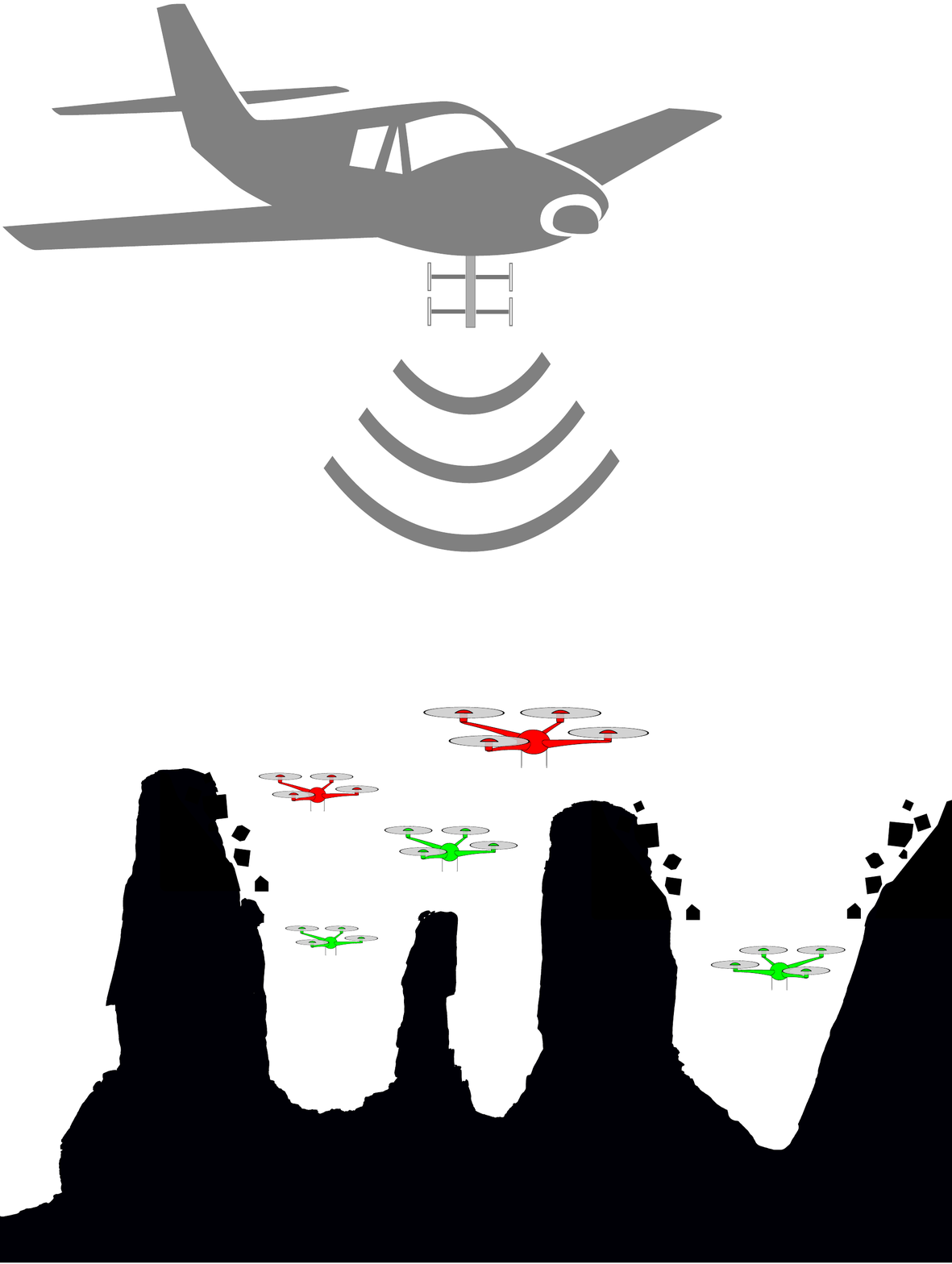}\\
	\caption{\textbf{Example scenario}}
\end{figure}

\section{Technical Approach}

The following section provides a technical description of the swarm mission planning and multi-object tracking techniques used in this paper.

\subsection{Swarm Mission Planning}

Central command must be able to autonomously determine the best mission plan for both pre-set and in-progress agent, target, and obstacle configurations. The technique chosen in this work is a coupled A*/Hungarian algorithm routine. 

A* is a pathfinding / graph traversal algorithm originally develeped in the 1960s \cite{astar} for the development of Shakey the Robot \cite{nilsson}. It achieves better performance than earlier methods such as Djikstra's algorithm by use of a heuristic to guide the search. This heuristic is set as the Euclidean distance between the subject node and the end node, the most common heuristic used. The cost of a node is then calculated as the number of nodes traversed to reach the subject node from the starting point plus the Euclidean distance from the subject node to the end point.

A* naturally handles obstacles potentially present during swarm operations. If a node is an obstacle, the algorithm will ignore it as a potential waypoint location. During real-life swarm operations these obstacles could be envisioned as buildings, known enemy locations, or environmental hazards such as terrain or wildfires. 

After paths are generated by A* for each agent/target combination, the optimal assignment algorithm determines which agent is assigned to which target. This is accomplished by selecting the assignment combination with the lowest total distance traveled by the team, using a method known as the Hungarian algorithm \cite{Kuhn_hungarian} \cite{munkres}. If the number of agents does not equal the number of targets, the algorithm will consider a different assignment metric. In this work only the situation with more targets than agents is considered, and agents are routed to their closest target while the targets farthest away from agents are disregarded. 

Once the optimal assignments are selected, the A* grid must be mapped to the physical world to generate spatial trajectories. A mesh is generated from the known mission area and A* grid, and each grid waypoint list supplied by A* is mapped to the physical world using that mesh. Grid size selection is a key mission design parameter. A fine grid will provide a more accurate path at the cost of computational effort, while a coarse grid will be easier computationally but will be less optimal in the real world and may not capture critical path details. The human mission initiator must carefully weigh these trade-offs based on the available computer hardware, expected obstacle density, and criticality of precise vehicle pathing.

Spatial trajectory waypoints are supplied to each agent and stored in local memory for use in the individual agent's guidance and control. Figures 3 demonstrates the STAT algorithm results from purely deterministic simulations with four agents/targets in an obstacle-dense environment. This shows that the underlying agent dynamics and swarm initial mission planning techniques are functional.

\begin{figure}\label{OneColumn}
	\centering
	\includegraphics[width=3.25in]{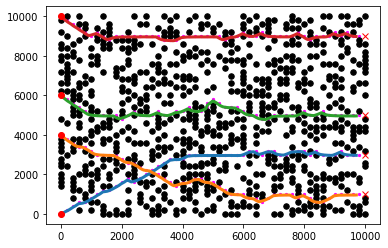}\\
	\caption{\textbf{Four agent/target simulation in obstacle-dense environment}}
\end{figure}

\subsection{Random Finite Set Multi-Object Tracking}
A Generalized Labeled Multi-Bernoulli (GLMB) random finite set (RFS) multi-object tracking technique is used for mission progress monitoring and swarm state estimation. A random finite set is a set of random vectors with unknown, time varying cardinality, or number of elements in the set. For the multi-object tracking case the set contains the state vectors for each object and the set's cardinality represents the number of tracked objects. The RFS multi-object tracking literature, with significant contributions by the Vo brothers, contains detailed technical descriptions of both the GLMB and predecessor methods [2-6]. A brief summary is provided here.

Because the elements of the set, here agent/target states, and the number of elements in the set are unknown, filters that rely on this formulation can also model object clutter, birth, spawn, and death. Here birth refers to new objects originating from some modeled probability distribution whereas object spawn refers to new objects created near existing objects. A real-world example of object spawn is an aircraft carrier launching fighters. The probabilistic nature of the RFS also allows the filters to model measurement clutter. The clutter model in this paper is a Poisson process using the rate provided to the RFS filter. A GLMB filter operates on a labeled RFS, which is simply an RFS that has distinct discrete labels appended to each state vector and there is a 1-1 mapping of states and labels.

Use of labeled multi-Bernoulli RFSs has been shown to outperform the Probability Hypothesis Density (PHD) \cite{Vo_GMPHD} and cardinalized PHD (CPHD) \cite{Vo_CPHD} due to its exploitation of both the strengths of the multi-Bernoulli RFS without the weaknesses of unlabeled MB RFS filters, which do not formally track estimates while exhibiting cardinality biases. \cite{ReuterLMB}

The Labeled Multi-Bernoulli filter theory was developed by B.T. Vo and B.N. Vo \cite{labeled_RFS_MO_Conj} and is summarized below. A Bernoulli RFS $X$ has a probability $1-r$ of being empty and a probability $r$ of being a set whose only element is distributed according to probability density $p$, otherwise known as the spatial distribution. The probability density for a Bernoulli RFS $X$ is given by

\begin{equation}
	\pi(X) = \begin{cases}
		1 - r & X = \emptyset \\
		r p(X) & X = \{x\}.
	\end{cases}
\end{equation}

A multi-Bernoulli RFS $X$ is then the union of $M$ independent Bernoulli RFSs and is completely defined by the parameter set $\{ (r^{i}, p^{i}) \}_{i=1}^{M}$. The probability density of a multi-Bernoulli RFS is described by

\begin{equation}
	\begin{split}
		\pi(\{x_1, \dotsc, x_n\}) &= \prod_{j=1}^{M}(1-r^{(j)}) \times \\
		&\sum_{1 \leq l_1 \neq \dotsi \neq i_n \leq M} \prod_{j=1}^{n}\frac{r^{(i_j)}p^{(i_j)(x_j)}}{1-r^{(i_j)}}\ .
	\end{split}
\end{equation}

In a GLMB each state $x \in$ $\mathcal{X}$ is augmented by a label $l \in$ $\mathcal{L}$ to establish the identity of a single object's trajectory. In implementation used for this work \cite{gasur} the labels are ordered pairs of integers where the first index is the time step of birth and the second index is a unique integer to distinguish targets born at the same time step. 

The GLMB filter must also solve the measurement association problem to properly propagate the state probability distributions. This is by accomplished by creating a hypothesis for each possible association (including clutter and agent births). Each hypothesis then has a measurement association probability, $w$, associated with it. These hypotheses also track the association history to generate a trajectory, and as such the total number of hypotheses grows with time. For practical use, the computations are kept tractable by pruning hypotheses with low probabilities. Finally, the density of a labeled multi-Bernoulli with parameter set $\{ (r^{l}, p^{l}) \}_{l \in L}$ is \begin{equation}
	\pi({\mathbf{X}}) = \Delta({\mathbf{X}}) w({\mathcal{L}}({\mathbf{X}})) p^{\mathbf{X}}
\end{equation} where the distinct label indicator $\Delta({\mathbf{X}}) = 1$ when the cardinality of ${\mathbf{X}}$ equals the cardinality of the labels. Additionally, the cardinality distribution of a labeled RFS is identical to the cardinality distribution of the unlabeled version. The hypothesis weights $w$ are \begin{align}
	w({\mathcal{L}}({\mathbf{X}})) &= \prod_{i\in {\mathbb{L}}}(1-r^{(i)}) \prod_{\ell \in {\mathbb{L}}} \frac{1_{\mathbb{L}}(\ell) r^{(\ell)}}{1-r^{\ell}} \\
	p^{\mathbf{X}} &= \prod_{\ell \in \mathbb{L}} p^{(\ell)}(x)\ .
\end{align} Note that the weights and spatial distributions satisfy normalization conditions, summing to one. Here $1_{\mathbb{L}}(\ell)$ is an indicator function, it is equal to one when $\ell \subset \mathbb{L}$ and zero otherwise.

One implementation of this filter is described in detail in \cite{VoLabeledRFS}, and only the main ideas will be summarized here. The filter handles bookkeeping and propagating/updating the hypothesis weights, as well as propagating/updating the underlying single object filters. These underlying single object filters are responsible for propagating and updating each object's state distribution, $p^{(\ell)}$, once the GLMB has determined a measurement association. This work closely follows the implementation described in \cite{VoLabeledRFS} which uses a Kalman filter to propagate individual states and represents each state probability, $p^{(\ell)}$, as a Gaussian mixture. Gaussian mixtures are simply a weighted sum of individual Gaussians. Equation 6 describes the probability density function of the Gaussian mixture, where $\mathbf{m}_i$ is the mean of the $i^{th}$ Gaussian, $P_i$ is the covariance of the $i^{th}$ Gaussian, and the weights $w_i$ sum to one. 


\begin{equation}
	f_{GM}(\mathbf{x}) = \sum_{i=1}^{N} w_i \mathcal{N} (\mathbf{x}; \mathbf{m}_i , P_i )
\end{equation}


This work utilizes two GLMB filters, one to track agent positions and one to track target positions. This eliminates the need to determine which filter outputs correspond with agents or targets. The outputs of the filter, in this work agent and target coordinates, are then provided to the swarm guidance update routine for consideration. So long as the agents and targets do not begin very close to one another each filter will be able to track the correct objects. This is because the birth locations of the agents are outside the birth probability distribution given to the target GLMB and therefore will be treated as clutter by that filter. The reverse is also true.

\section{Simulation Description}

The following section describes the simulation methodology. This work utilizes a linear state space model based on the lateral dynamics of a small fixed-wing aircraft for the agent motion. Body-frame forward velocity $u$ is assumed to be regulated by a longitudinal control system and is pre-set by the human mission initiator. Agent locations are propagated by numerically integrating the body velocities transformed to the global frame via the heading angle. The agent dynamics model is described in the conference paper \cite{aero2021}.

The agent dynamics are controlled by a full state feedback linear-quadratic-regulator (LQR) controller with $\underline{Q} = \underline{I}_5$ and $\underline{R} = \underline{I}_2$. The augmented system is discretized at 100 Hz for simulation. The heading command is determined from the angle between the agent's current position and the active waypoint. When the Euclidean distance between the agent and its current waypoint reaches a certain threshold, in this work 10 meters, the active waypoint updates to the next in the list supplied by central command and mapped to the continuous physical domain. Target motion is via a 2D double integrator dynamic model discretized at 100 Hz for simulation. $\dot{x}$ and $\dot{y}$ are either pre-set or determined via a Gaussian distribution in this work. Agent death occurs via a seeded random process, with dead agents simply no longer propagating dynamics or producing measurements.

The formulation described in Section 2 was implemented in Python in conjunction with two public code packages developed by the authors, GNC for Autonomous Swarms Using RFS (GASUR) \cite{gasur} and GNCPy \cite{gncpy}, and used to generate a number of example cases. GASUR was used for its GLMB implementation, while GNCPy was used for its Kalman filter classes.

The techniques described above were cohered in a single code package to perform numerical simulations. The algorithm begins by initializing the agents, targets, and mission area and performing an initial run of the A* algorithm based on the modeled agent/target starting locations. The starting points of the pathfinding algorithm are the agent locations, while the end points are the target locations. Then the main simulation loop begins, with agent/target dynamics propagating at 100 Hz, the clutter process and GLMB filters operating at 1 Hz, and the swarm guidance updating and agent death process occurring at 0.2 Hz. If the RFS filter outputs indicate that a swarm guidance update is necessary, the starting points of the pathfinding algorithm are set to the outputs of the agent GLMB and the ending points are set to the outputs of the target GLMB. Once all of the agents have either been disabled or reached a target, the simulation ends. Algorithm 1 in the appendix provides a description of the implementation in this work.

Several mission simulations, each with four agents and five targets, are presented below. Agents begin each mission on the left side of the map, while targets begin on the right.  Obstacles are generated by both a uniformly distributed random number generator and hand placement by the simulation operator. For the random obstacle cases, a uniform random number is generated between 0 and 1 for each A* node. If the number is less than a prescribed threshold the node is chosen as an obstacle. A* nodes with an obstacle present are off-limits for the pathfinding algorithm. Note that targets are allowed to overfly obstacles to present more of a challenge to the swarm guidance computation and that obstacles do not generate clutter measurements.

\section{Simulation Results}

Figures 4 and 5 demonstrate results of a simulation in which target velocity and starting locations are prescribed. Figure 4 shows the true agent and target motion with red o's and x's as initial and final positions, respectively, and obstacles as black dots. The first plot in Figure 4 is a mid-mission snapshot, while the second shows the end of the simulation. Figure 5 plots the end-mission states for both the agent and target GLMB RFS filters in black-outlined circles and clutter measurements as transparent gray triangles. It is clear that when the agent represented by the green trace in Figure 4 is disabled, the orange trace agent is rerouted to the yellow trace target, since that target is now closest to that agent. These results show that the GLMB filter tracks the agent and target states and that the GLMB outputs can be used to update the swarm guidance during mission execution. Waviness in the agent paths is due to the 0.2 Hz A* update rate. This can be ameliorated by increasing that rate, but a compromise must be made between performance and computational requirements. Occasionally, clutter measurements that happen to appear in close proximity to dead agents will be processed as states. These can be seen in the GLMB output figures as single dots near the termination of a disabled agent's trajectory. This occurs because the GLMB filter does not know that the agent is dead, and interprets new measurements in the disabled agent's vicinity as true agent motion. The filter then believes that the single clutter measurement is that dead agent and produces a spurious state at that time step. This has no affect on mission accomplishment because the spurious state is only present for one time step, usually far from the surviving agents. Even if it does cause an agent to re-route to a different target this will be corrected at the next A* update, in this work after only five seconds.

\begin{figure}\label{OneColumn}
	\centering
	\includegraphics[width=3.25in]{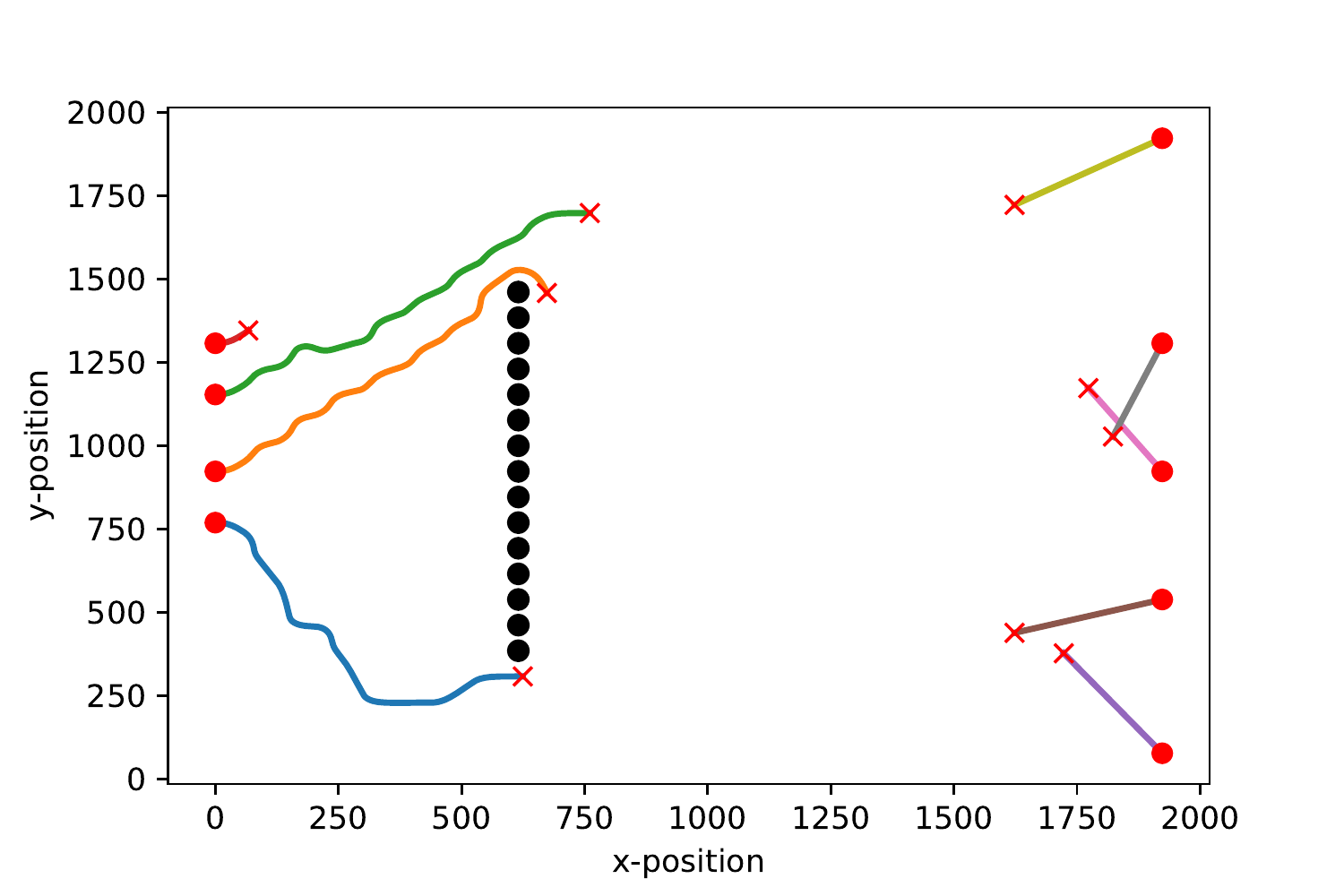}\\
	\includegraphics[width=3.25in]{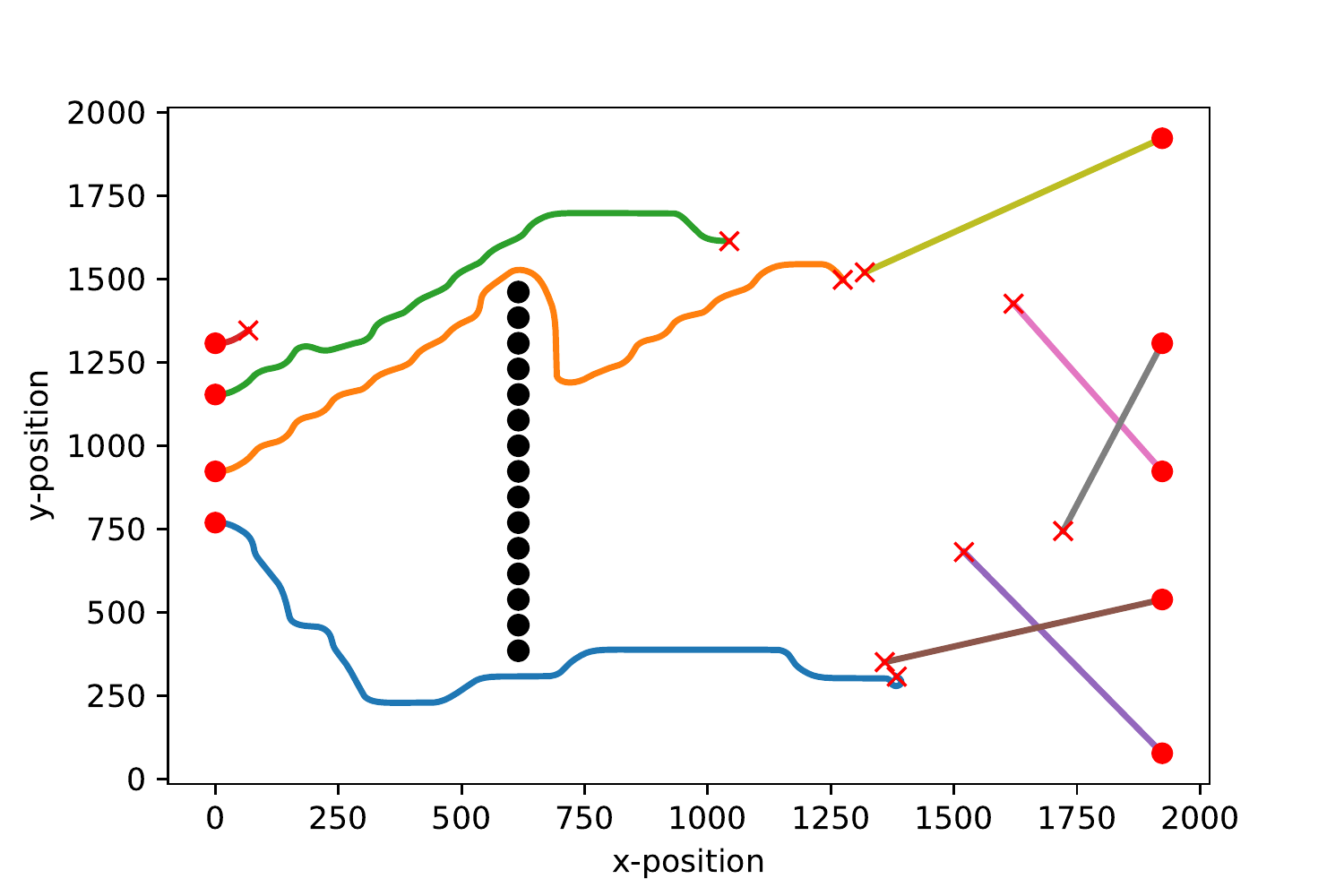}\\
	\caption{\textbf{Ground truth for prescribed obstacle locations}}
\end{figure}

\begin{figure}\label{OneColumn}
	\centering
	\includegraphics[width=3.25in]{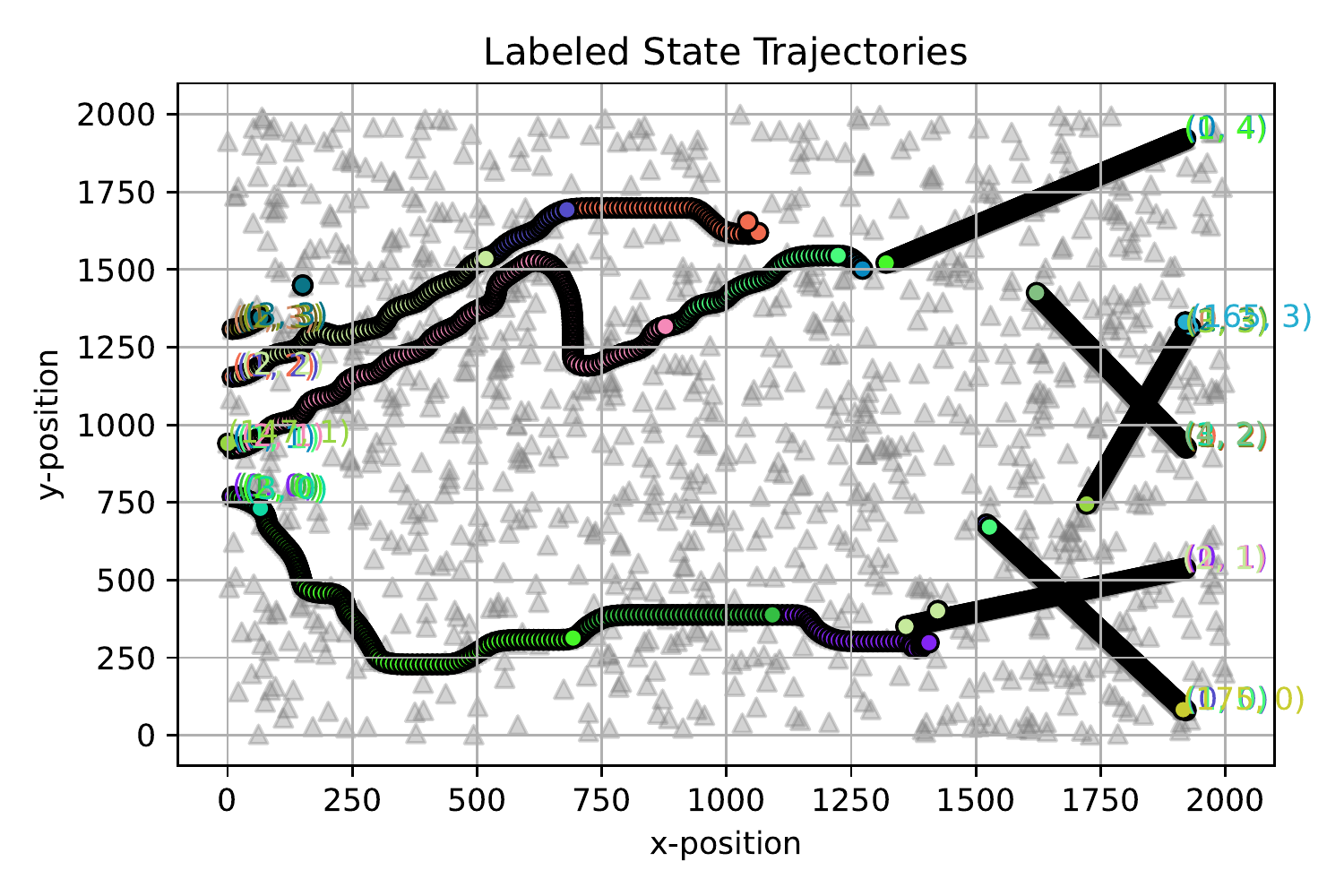}\\
	\caption{\textbf{GLMB output for prescribed obstacle locations}}
\end{figure}

A more interesting and appropriate challenge for this routine is a mission area with randomly placed obstacles. Figures 6 and 7 present the results of this simulation. Figure 6 presents the mid and end-mission ground truth, while Figure 7 demonstrates the robustness of the GLMB to very cluttered measurements and partially overlapping object trajectories. It is clear that this technique can handle high obstacle density with targets moving freely in the mission area. Another similar run was performed with results in Figure 8 and 9. In this simulation target motion and birth locations are randomly assigned via Gaussian distributions. The results in Figures 6-9 support the claim that this technique is able to generate plans for the swarm to complete chaotic missions with high obstacle density, heavily cluttered measurements, agent death, and complicated target behavior. In each simulation the mission plan was altered to account for agent death and target motion, and did not allow measurement clutter to affect the mission. More extensive simulations will be explored as the architecture design is improved.

\begin{figure}\label{OneColumn}
	\centering
	\includegraphics[width=3.25in]{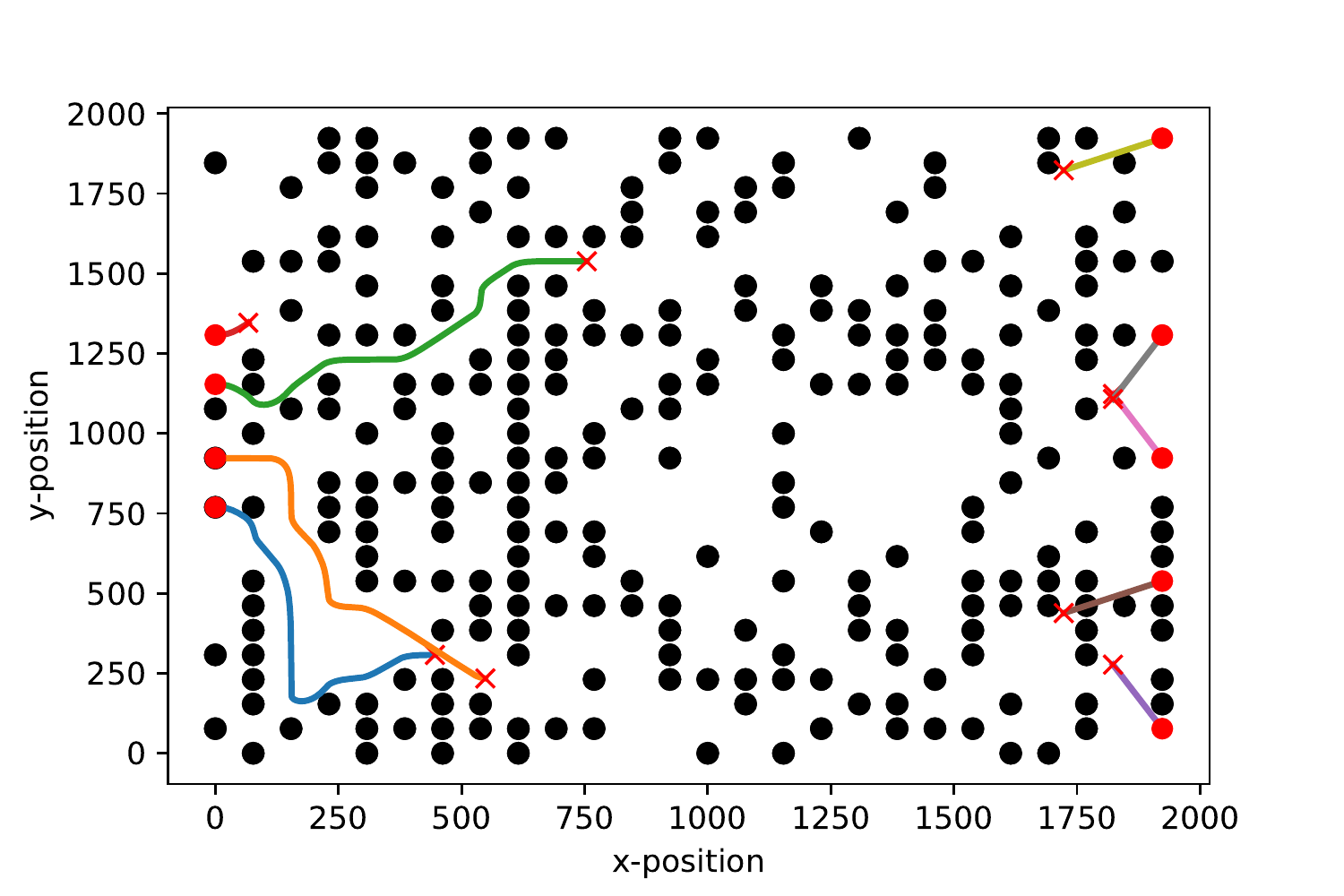}\\
	\includegraphics[width=3.25in]{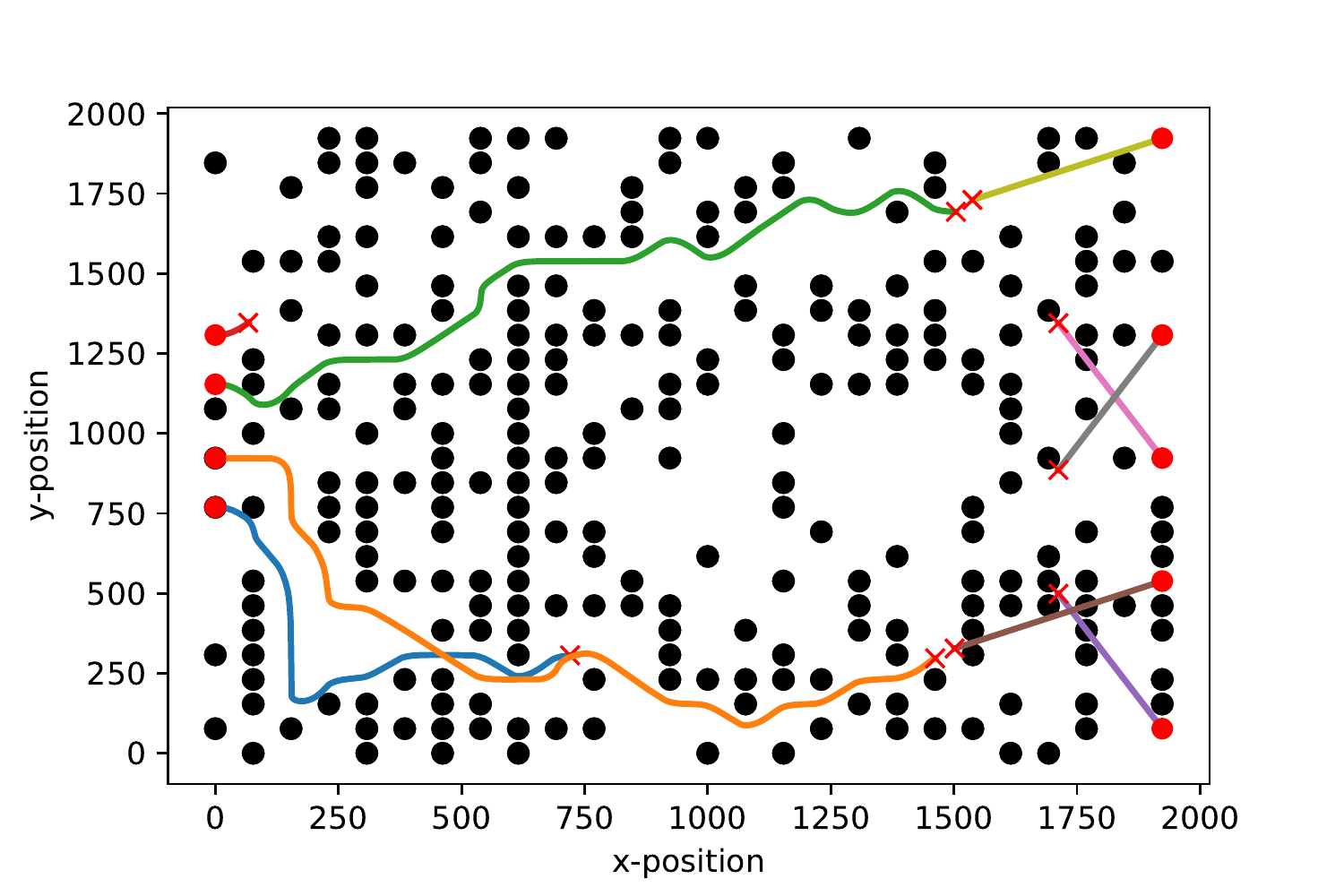}\\
	\caption{\textbf{Ground truth for random obstacle locations}}
\end{figure} 

\begin{figure}\label{OneColumn}
	\centering
	\includegraphics[width=3.25in]{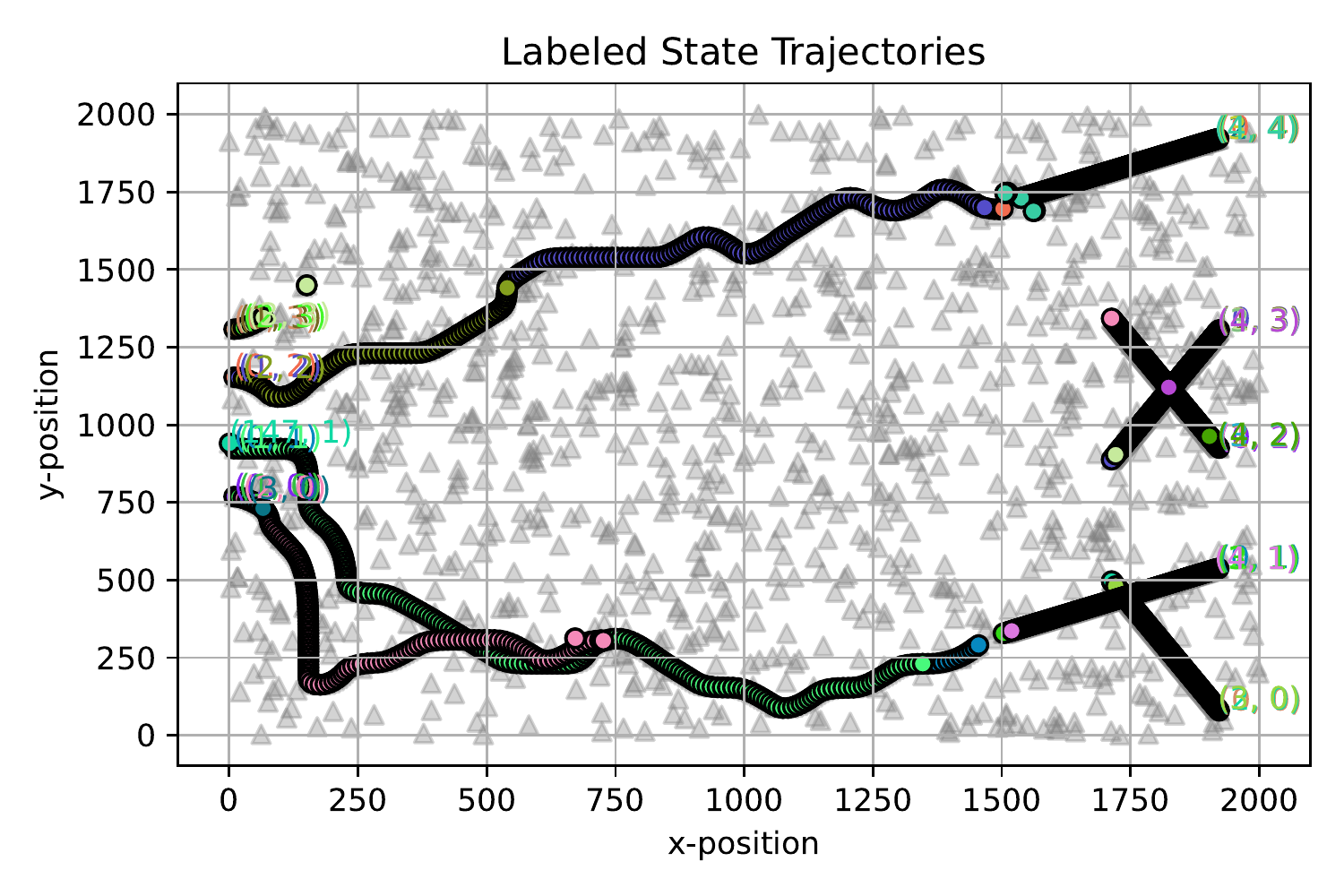}\\
	\caption{\textbf{GLMB output for random obstacle locations}}
\end{figure}

\begin{figure}\label{OneColumn}
	\centering
	\includegraphics[width=3.25in]{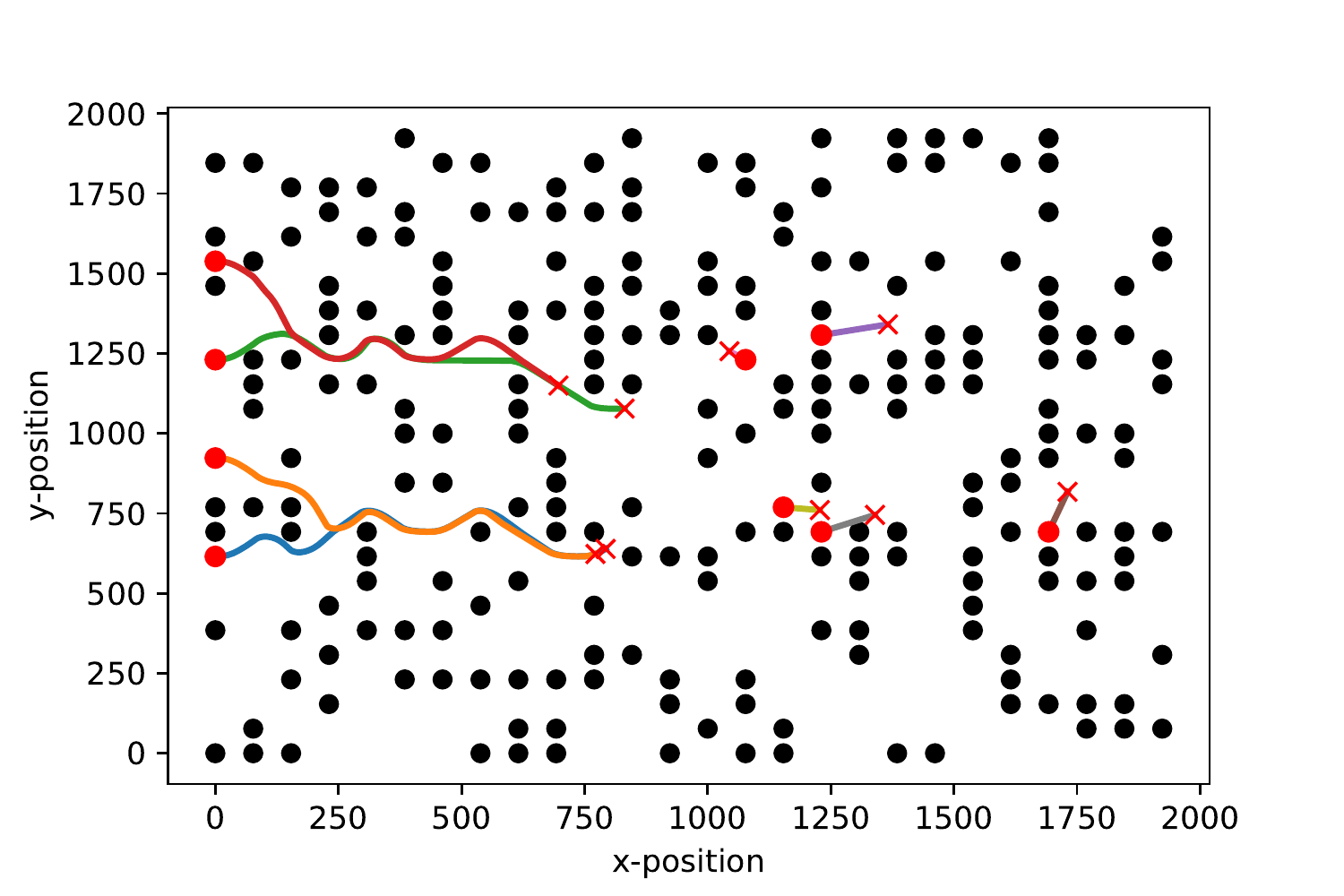}\\
	\includegraphics[width=3.25in]{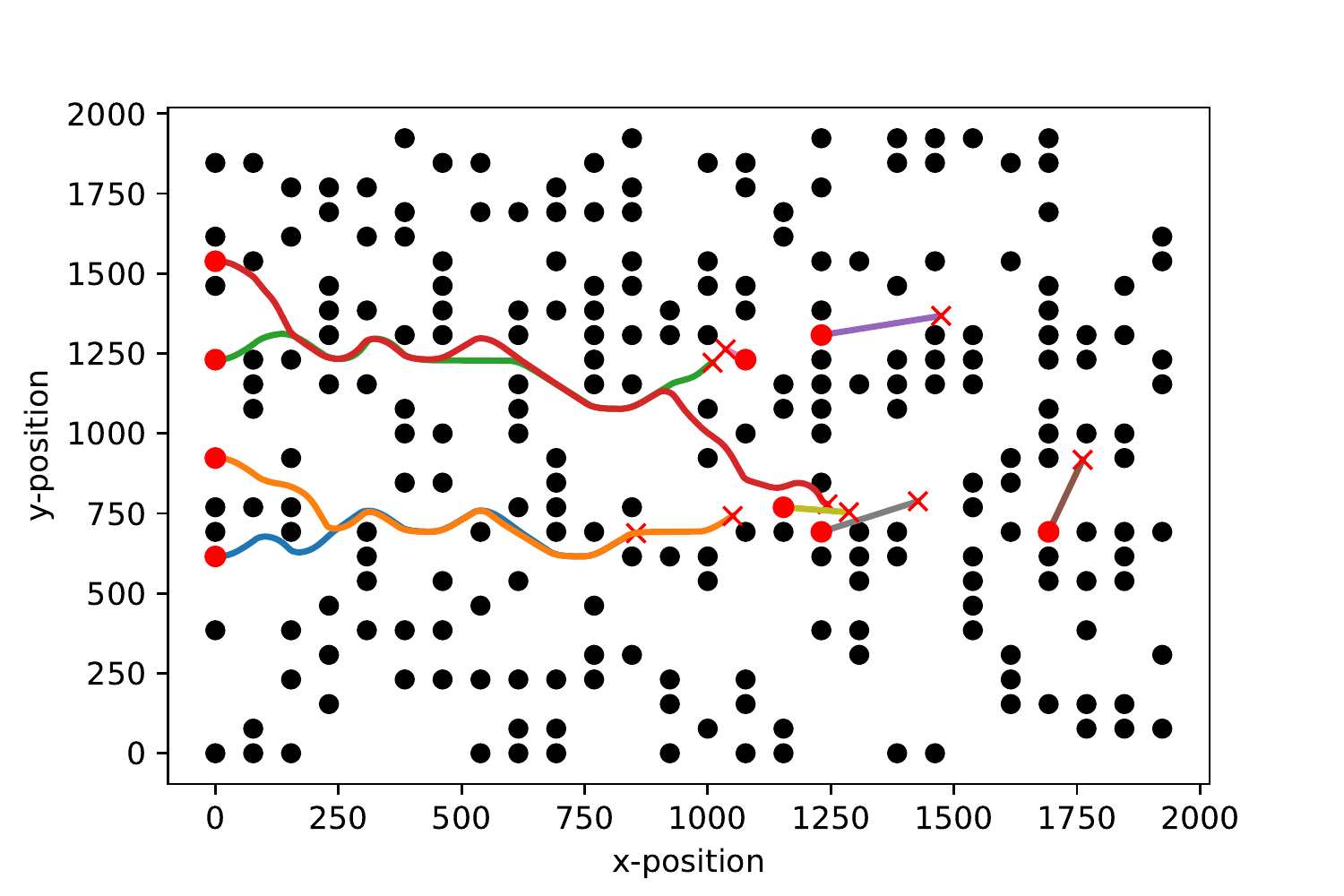}\\
	\caption{\textbf{Ground truth for random target location/motion}}
\end{figure}

\begin{figure}\label{OneColumn}
	\centering
	\includegraphics[width=3.25in]{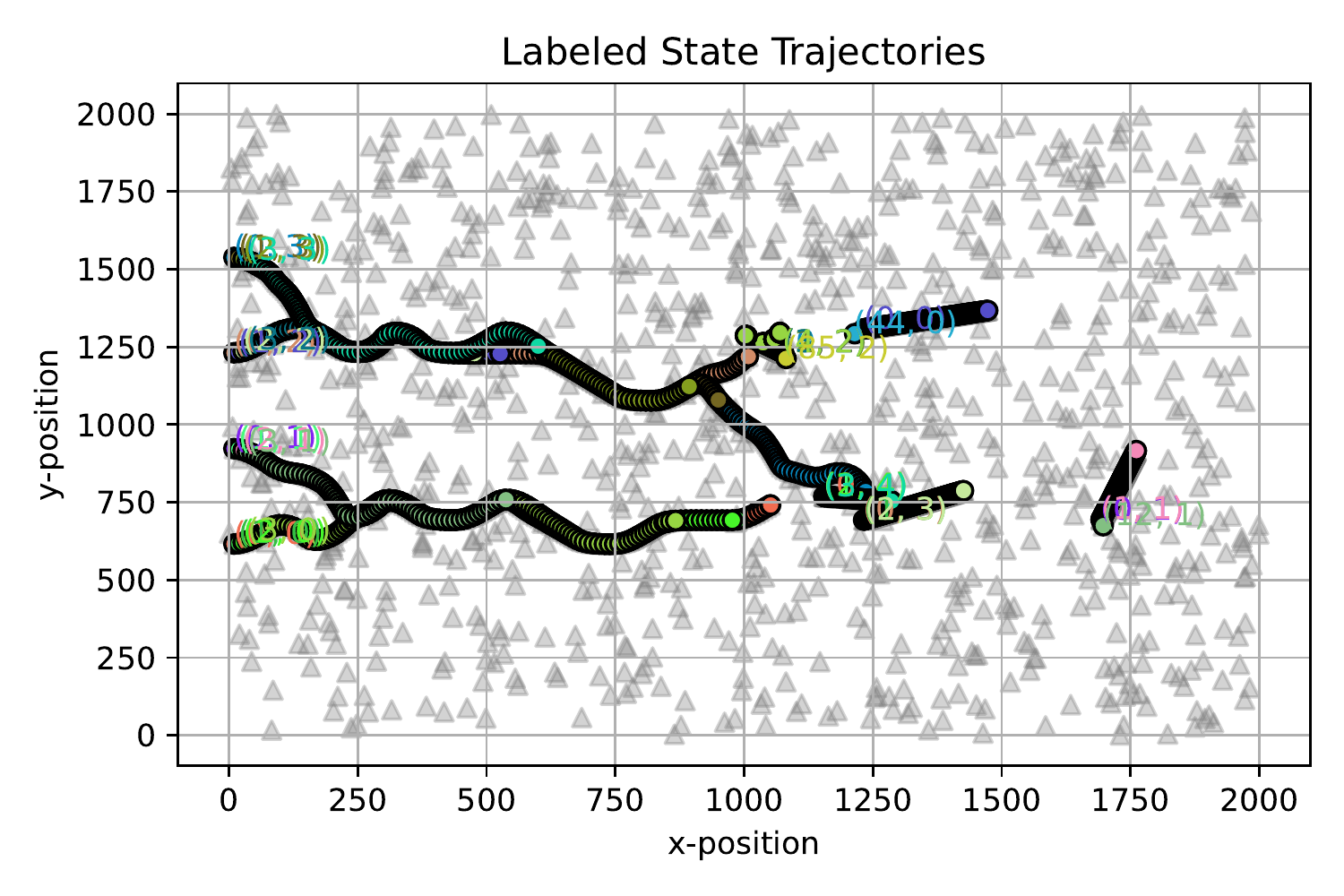}\\
	\caption{\textbf{GLMB output for random target location/motion}}
\end{figure}


\section{Conclusions}

A technique for autonomous robotic swarm dynamic mission planning was presented in this work. Simulations showed that the algorithm is robust to agent death, target motion, numerous obstacles in the mission area, and highly cluttered measurements.

This method can be used for arbitrarily many agents, but the computational requirements on consumer hardware become burdensome as cardinality increases due to the computational stress of the GLMB and $\mathcal{O}$$(n^{2})$ complexity in the A* loop. A recent development in RFS MOT technology using joint prediction and update steps for the GLMB \cite{jointglmb} can be implemented in the GASUR code package to improve efficiency. As well as RFS filter improvements, use of a sample-based pathfinding algorithm such as Rapidly exploring Random Tree* (RRT*) \cite{rrtstar1} has potential for computational effort reduction. RRT* in particular has been applied with motion constraints that guarantee the subject vehicle can execute the path \cite{rrt} \cite{rrt2}. Additionally, the CPU used in these simulations was also burdened by 100 Hz dynamics updates and is of consumer grade. In applications central command could use a dedicated high-quality CPU that would only have to perform the GLMB filtering and A* swarm guidance updates to greatly increase execution speed.

A related issue is the lack of target motion prediction. The current technique is not forward-looking, instead relying on target measurements to update the swarm guidance. Targets from this paper only move at approximately half the speed of the agents, but for targets moving at or above the agent velocity target state prediction is necessary because otherwise targets would be able to outrun agents. This problem is addressed in a work by the second and third authors \cite{JDL_Thomas_IRL} that uses coupled Deep Q-learning and Inverse Reinforcement learning to estimate target dynamics purely from the GLMB output and predict object motion over a given time horizon. Modifying a more simple target intercept technique such as proportional navigation guidance for use with A* is promising. Combining intercept functionality with the techniques in this paper would allow mission completion with faster targets, and also improve swarm performance with slower targets by setting the final waypoint at the predicted location rather than "chasing" the targets in the current scheme. These issues present an impediment for real-world applications where quick responses to fast-moving targets are crucial. 


The results demonstrate good initial performance and promise for further development. Topics of interest to the authors include decentralized operations, target motion prediction, and sophisticated guidance techniques.

\appendix

\begin{algorithm}[H]
	\caption{Simulation Algorithm}
	\begin{algorithmic}[1]
		\STATE Initialize agents, targets, and mission region
		\STATE Generate A* map
		\FOR {each agent/target combination}
		\STATE Initial A* search
		\ENDFOR
		\STATE Optimal target assignment
		\STATE Initialize GLMB filter
		\\ \textbf{Main simulation loop:}
		\\ \textbf{At 100 Hz:} 
		\FOR {each agent}
		\STATE Propagate dynamics and position
		\STATE Generate measurements
		\STATE Check for waypoint/mission completion
		\ENDFOR
		\FOR {each target}
		\STATE Propagate dynamics and position
		\STATE Generate measurements
		\ENDFOR
		\\ \textbf{At 1 Hz:} 
		\STATE Generate clutter measurements
		\STATE Run GLMB filters
		\FOR {each agent}
		\STATE Run death process
		\ENDFOR
		\\ \textbf{At 0.2 Hz:}
		\FOR {each agent}
		\STATE Check if agent has died
		\IF{true}
		\STATE flag = 1
		\ENDIF
		\ENDFOR
		\FOR {each target}
		\STATE Check if movement greater than threshold
		\IF{true}
		\STATE flag = 1
		\ENDIF
		\ENDFOR
		\IF {flag == 1}		
		\STATE Run A* search 
		\ENDIF
		\STATE continue
	\end{algorithmic}
\end{algorithm}

\bibliographystyle{ieeecolor}
\bibliography{\myreferences}

\begin{thebibliography}{10}
\providecommand{\url}[1]{#1}
\csname url@samestyle\endcsname
\providecommand{\newblock}{\relax}
\providecommand{\bibinfo}[2]{#2}
\providecommand{\BIBentrySTDinterwordspacing}{\spaceskip=0pt\relax}
\providecommand{\BIBentryALTinterwordstretchfactor}{4}
\providecommand{\BIBentryALTinterwordspacing}{\spaceskip=\fontdimen2\font plus
\BIBentryALTinterwordstretchfactor\fontdimen3\font minus
\fontdimen4\font\relax}
\providecommand{\BIBforeignlanguage}[2]{{%
\expandafter\ifx\csname l@#1\endcsname\relax
\typeout{** WARNING: IEEEtran.bst: No hyphenation pattern has been}%
\typeout{** loaded for the language `#1'. Using the pattern for}%
\typeout{** the default language instead.}%
\else
\language=\csname l@#1\endcsname
\fi
#2}}
\providecommand{\BIBdecl}{\relax}
\BIBdecl

\bibitem{cluttrack}
A.~Pierson and D.~Rus, ''Distributed target tracking in cluttered environments with guaranteed collision avoidance,'' 2017 International Symposium on Multi-Robot and Multi-Agent Systems (MRS), 2017, pp. 83-89, doi: 10.1109/MRS.2017.8250935.

\bibitem{DOC}
G.~Foderaro, P.~Zhu, H.~Wei, T.~A. Wettergren and S.~Ferrari, ''Distributed Optimal Control of Sensor Networks for Dynamic Target Tracking,'' in IEEE Transactions on Control of Network Systems, vol.~5, no.~1, pp.~142-153,~March 2018,~doi:~10.1109/TCNS.2016.2583070.

\bibitem{coop_sense}
K.~You, Q.~Chen, P.~Xie and S.~Song, "Range-Based Coordinate Alignment for Cooperative Mobile Sensor Network Localization," in IEEE Transactions on Control of Network Systems, vol. 7, no. 3, pp. 1379-1390, Sept. 2020, doi: 10.1109/TCNS.2020.2977334.

\bibitem{Mahler_MTBayes}
R.~Mahler, ``Multi-target Bayes filtering via first-order
multi-target moments,'' \emph{IEEE Transactions on AES}, 
vol.~39, no.~4, pp. 1152--1178, 2003.

\bibitem{Vo_GMPHD}
B.-N. Vo and W.-K. Ma, ``The gaussian mixture probability hypothesis density
filter,'' \emph{IEEE Transactions on signal processing}, vol.~54, no.~11, pp.
4091--4104, 2006.

\bibitem{Vo_CPHD}
B.-T. Vo, B.-N. Vo, and A.~Cantoni, ``The cardinalized probability hypothesis
density filter for linear gaussian multi-target models,'' in \emph{2006 40th
Annual Conference on Information Sciences and Systems}.\hskip 1em plus 0.5em
minus 0.4em\relax IEEE, 2006, pp. 681--686.

\bibitem{labeled_RFS_MO_Conj}
B.-T. Vo and B.-N. Vo, ``Labeled random finite sets and multi-object conjugate
priors,'' \emph{IEEE Transactions on Signal Processing}, vol.~61, no.~13, pp.
3460--3475, Jul 2013.

\bibitem{ReuterLMB}
S.~Reuter, B.-T. Vo, B.-N. Vo, and K.~Dietmayer, ``The labeled multi-bernoulli
filter,'' \emph{IEEE Transactions on Signal Processing}, vol.~62, no.~12, pp.
3246--3260, 2014.

\bibitem{VoLabeledRFS}
B.-T. Vo, B.-N. Vo, and D.~Phung, ``Labeled random finite sets and the bayes
  multi-target tracking filter,'' \emph{IEEE Transactions on Signal
  Processing}, vol.~62, no.~24, pp. 6554--6567, Dec 2014.

\bibitem{jointglmb}
H. G. Hoang, B. T. Vo and B. Vo, "A fast implementation of the generalized labeled multi-Bernoulli filter with joint prediction and update," 2015 18th International Conference on Information Fusion (Fusion), 2015, pp. 999-1006.

\bibitem{doerr_conf}
B. Doerr and R. Linares, "Control of Large Swarms via Random Finite Set Theory," 2018 Annual American Control Conference (ACC), 2018, pp. 2904-2909, doi: 10.23919/ACC.2018.8430968.

\bibitem{doerr_arxiv}
B. Doerr and R. Linares. “Decentralized Control of Large Collaborative Swarms Using Random Finite Set Theory.” ArXiv:2003.07221 [Cs, Eess], Mar. 2020. arXiv.org, http://arxiv.org/abs/2003.07221

\bibitem{astar}
P.E.~Hart, N.J.~Nilsson, and B.~Raphael, ``A Formal Basis for the Heuristic Determination 
of Minimum Cost Paths,'' \emph{IEEE Transactions on Systems Science and Cybernetics,}, 
vol.~39, pp. 100--107, 1968.

\bibitem{nilsson}
N.J.~Nilsson, ``The Quest for Artificial Intelligence,'' \emph{Cambridge University Press,}, 
2010.

\bibitem{Kuhn_hungarian}
H.W.~Kuhn, ``The Hungarian Method for the Assignment Problem,'' \emph{Naval Research Logistics Quarterly,} 
pp 83-87, 1955.

\bibitem{munkres}
J.~Munkres, ``Algorithms for the Assignment and Transporation Problems,'' \emph{Journal of the Society for 
Industrial and Applied Mathematics,} pp 32-38, 1957.

\bibitem{JDL_Linares_ION}
J.D.~Larson, B.~Doerr, and R.~Linares, ``Autonomous
Mission Planning for Swarms Using Random Finite
Sets,'' \emph{Proceedings of the 32nd International Technical
Meeting of the Satellite Division of The Institute of
Navigation (ION GNSS+ 2019)} pp 1753-1761, 2019.

\bibitem{JDL_Thomas_ELQR}
R. W. Thomas and J. D. Larson, "Receding Horizon Extended Linear Quadratic Regulator for RFS-based Swarms with Target Planning and Automatic Cost Function Scaling," in IEEE Transactions on Control of Network Systems, doi: 10.1109/TCNS.2021.3050133.

\bibitem{hicard}
R.W. Thomas, V. W. Hill, and J. D. Larson, “Hierarchical GNC for High Cardinality Random Finite Set Based Teams with Autonomous Mission Planning.” AIAA Scitech 2021 Forum, American Institute of Aeronautics and Astronautics, 2021. arc.aiaa.org (Atypon), doi:10.2514/6.2021-0268.

\bibitem{JDL_Thomas_IRL}
R.W.~Thomas and J.D.~Larson, ``Inverse Reinforcement Learning for Generalized Labeled Multi-Bernoulli
Multi-Target Tracking,'' \emph{IEEE Aerospace 2021,} to be published.

\bibitem{aero2021}
V. W. Hill, R. W. Thomas, and J. D. Larson, “Autonomous Situational Awareness for UAS Swarms.” ArXiv:2104.08904 [Cs, Eess], Apr. 2021. arXiv.org, http://arxiv.org/abs/2104.08904.

\bibitem{gasur}
\BIBentryALTinterwordspacing
J.~D. Larson, R.~W. Thomas, V.~W. Hill, and V.~Weirens, ``{GASUR}: A {P}ython library for
{G}uidance, navigation, and control of {A}utonomous {S}warms {U}sing {R}andom
finite sets,'' Web page, 2019. [Online]. Available:
\url{https://github.com/drjdlarson/gasur}
\BIBentrySTDinterwordspacing

\bibitem{gncpy}
\BIBentryALTinterwordspacing
J.~D. Larson and R.~W. Thomas, ``{GNCPy}: A {P}ython library for {G}uidance,
{N}avigation, and {C}ontrol algorithms,'' Web page, 2019. [Online].
Available: \url{https://github.com/drjdlarson/gncpy}
\BIBentrySTDinterwordspacing

\bibitem{rrt}
LaValle, S.,''Rapidly-exploring random trees : a new tool for path planning,'' The annual research report, 1998.

\bibitem{rrt2}
LaValle SM, Kuffner JJ., ''Randomized Kinodynamic Planning''. The International Journal of Robotics Research. 2001;20(5):378-400. doi:10.1177/02783640122067453


\bibitem{rrtstar1}
Karaman S, Frazzoli E. Sampling-based algorithms for optimal motion planning. The International Journal of Robotics Research. 2011;30(7):846-894. doi:10.1177/0278364911406761


\end{thebibliography}

%
%

\begin{IEEEbiography}[{\includegraphics[width=1in,height=1.25in,clip,keepaspectratio]{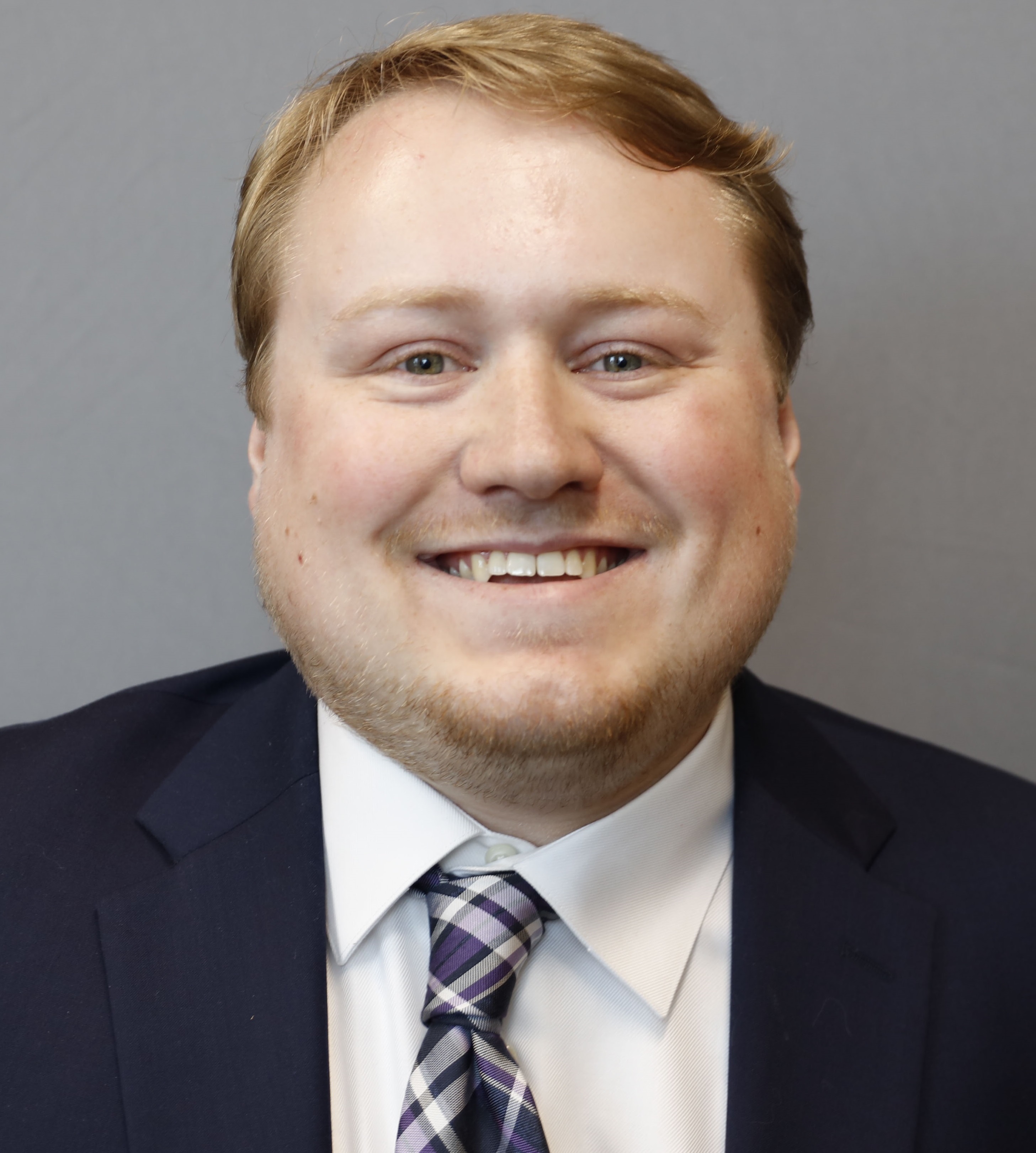}}]{Vincent W. Hill}  
received the B.S. degree in mechanical engineering and the M.S. degree in aerospace engineering and mechanics from the University of Alabama, Tuscaloosa, AL, USA in 2017 and 2020, respectively. His research interests include autonomous systems, multi-vehicle GNC, and flexible structure control. In parallel with pursuing his doctorate he is currently employed with Aerovironment, inc. in Simi Valley, CA as a GNC Engineer on the solar high-altitude long-endurance aircraft program.
\end{IEEEbiography}

\begin{IEEEbiography}[{\includegraphics[width=1in,height=1.25in,clip,keepaspectratio]{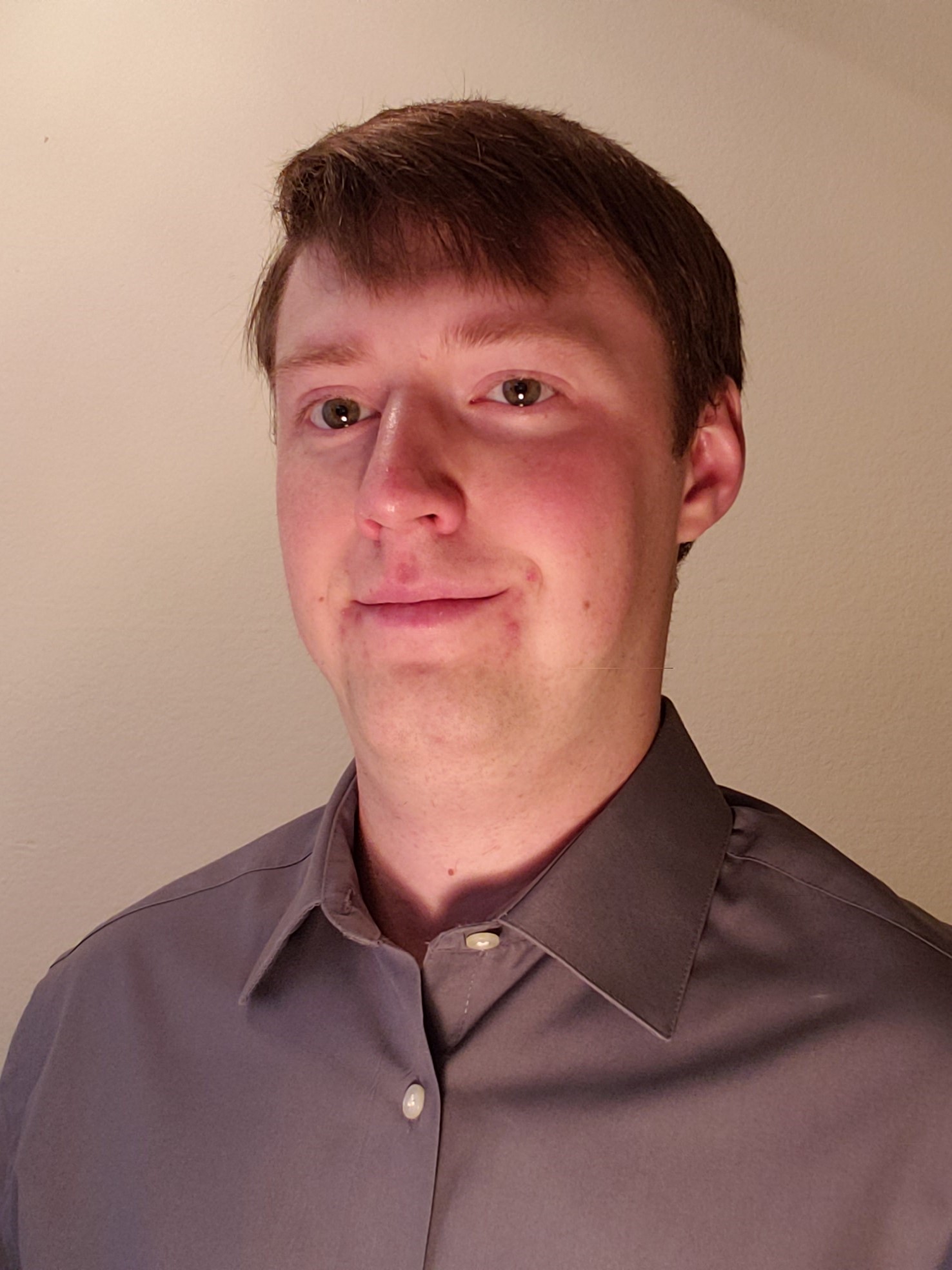}}]{Ryan W. Thomas}  
	received a B.S. degree in aerospace engineering from the University of Minnesota, Minneapolis, MN, USA in 2019. His research interests include autonomous systems, guidance, navigation, and control, and multi-agent/multi-sensor systems.
\end{IEEEbiography}

\begin{IEEEbiography}[{\includegraphics[width=1in,height=1.25in,clip,keepaspectratio]{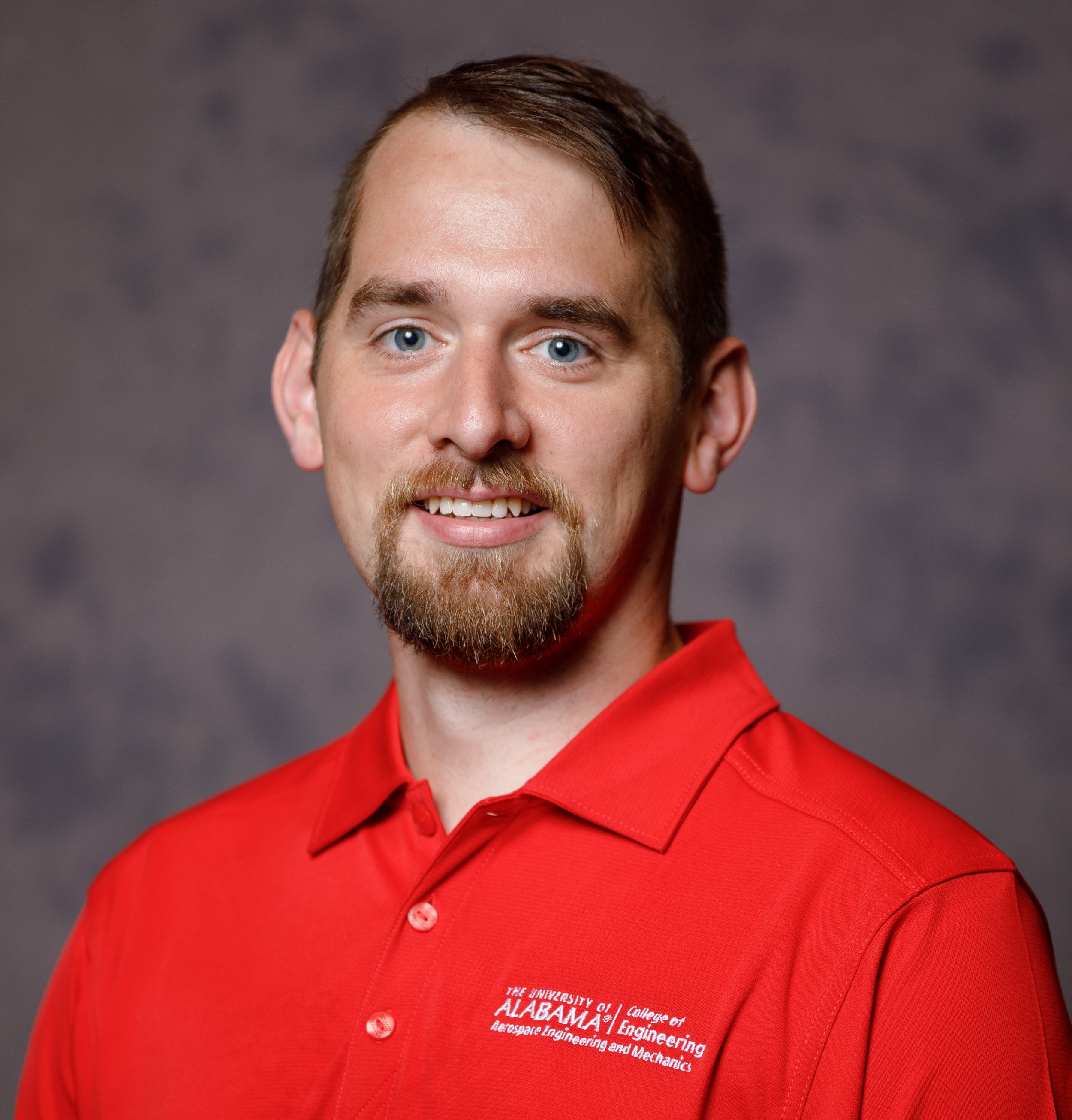}}]{Jordan D. Larson} 	
received the B.S., M.S., and Ph.D. degrees in aerospace engineering from University of Minnesota, Minneapolis, MN, USA in 2013, 2015, and 2018, respectively. From 2018 to 2019, he was a Post Doctoral Research Associate with the University of Minnesota before becoming an assistant professor at the University of Alabama. His research interests include autonomous systems, guidance, navigation, and control, multi-agent/multi-sensor systems, signal processing, and uncertainty quantification. Dr. Larson also recently received the Samuel M. Burka Award from the Institute of Navigation. 
\end{IEEEbiography}
	
\end{document}